\documentclass[lettersize,journal]{IEEEtran}
\usepackage{booktabs}
\usepackage{array}
\usepackage[caption=false,font=normalsize,labelfont=sf,textfont=sf]{subfig}
\usepackage{cite}
\usepackage{amsmath,amssymb,amsfonts}
\usepackage{algorithmic}
\usepackage{algorithm}
\usepackage{graphicx}
\usepackage{textcomp}
\usepackage{xcolor}
\usepackage{varwidth}
\usepackage{stfloats}
\usepackage{url}
\usepackage{verbatim}
\usepackage{tabu}                   
\usepackage{multirow}                
\usepackage{multicol}                
\usepackage{multirow}               
\usepackage{float}                   
\usepackage{makecell}                
\usepackage{booktabs}  
\usepackage{hyperref}

\def\BibTeX{{\rm B\kern-.05em{\sc i\kern-.025em b}\kern-.08em
    T\kern-.1667em\lower.7ex\hbox{E}\kern-.125emX}}

\title{CrackESS: A Self-Prompting Crack Segmentation System for Edge Devices}

\author{
	Yingchu Wang\IEEEauthorrefmark{1}, 
	Ji He\IEEEauthorrefmark{2}, 
	and Shijie Yu\IEEEauthorrefmark{3}
}

\begin{document}
	\maketitle
	
\begin{abstract}
	Structural Health Monitoring (SHM) is a sustainable and essential approach for infrastructure maintenance, enabling the early detection of structural defects. Leveraging computer vision (CV) methods for automated infrastructure monitoring can significantly enhance monitoring efficiency and precision. However, these methods often face challenges in efficiency and accuracy, particularly in complex environments. Recent CNN-based and SAM-based approaches have demonstrated excellent performance in crack segmentation, but their high computational demands limit their applicability on edge devices. This paper introduces CrackESS, a novel system for detecting and segmenting concrete cracks. The approach first utilizes a YOLOv8 model for self-prompting and a LoRA-based fine-tuned SAM model for crack segmentation, followed by refining the segmentation masks through the proposed Crack Mask Refinement Module (CMRM). We conduct experiments on three datasets(Khanhha's dataset, Crack500, CrackCR) and validate CrackESS on a climbing robot system to demonstrate the advantage and effectiveness of our approach.
\end{abstract}


\section{Introduction}
In the field of defect prevention, structural health monitoring (SHM) is a crucial method for protecting infrastructure such as roads and bridges by detecting potential structural collapse and physical damage \cite{1}. To overcome the high costs and low efficiency of traditional SHM methods \cite{2}, an increasing number of advanced computer vision-based technologies have been developed and implemented in automatic assessment. 

The application of computer vision techniques to SHM, particularly for defects pixel-level detection (segmentation), has been researched for decades and classified in two branches. The traditional image processing approaches utilize histograms, morphology, filtering and model analysis \cite{3} to identify cracks, such as filtering method based approach \cite{4}. The deep learning-based approaches have designed various CNN architectures for crack segmentation, such as CrackNet \cite{5}, which removes max pooling layers to improve accuracy, and U-Net based networks \cite{6} and \cite{7} , which leverage multi-level feature fusion.  In addition, some efforts have focused on a foundation model, Segment Anything Model (SAM), to enable the method to have semantic understanding and zero-shot capability simultaneously. The methods have been demonstrated effectiveness in the detection of defects, \cite{8} and \cite{9} propose fine-tuned SAM models by using Low-Rank Adaptation (LoRA) for crack segmentation, and they have achieved remarkable performance in experiments conducted on open-source datasets. Similar to the significant impact SAM has had in fields, such as medical imaging, agriculture, and manufacturing, the fine-tuned SAM model offers substantial advantages for structural health monitoring \cite{10}.
\begin{figure}[!t]
	\centering
	\includegraphics[width=3.5in]{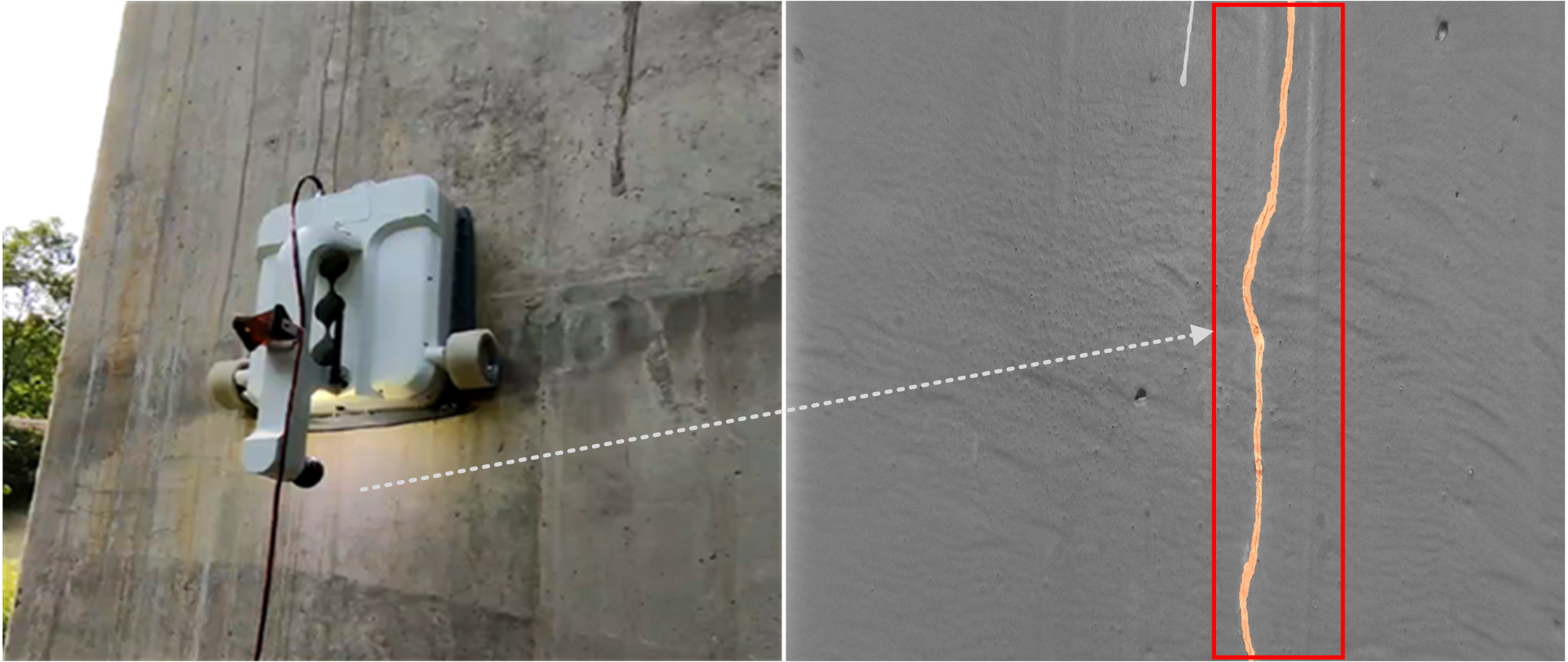}
	\caption{(a) The climbing robot executed an inspection task on a concrete bridge pier. (b) The result of inspection }
	\label{fig1}
\end{figure}

However, existing SAM-based crack detection methods face the challenge that the learning step and computational resource consumption are expensive, making it difficult to deploy in real-world engineering applications, especially on low-cost autonomous platforms such as climbing robot systems \cite{11} and UAV monitoring systems \cite{12}. Moreover, high-resolution industrial cameras are often used on robotic platforms to detect fine cracks. Therefore, developing a lightweight, accurate, high-resolution adaptable crack detection system that is compatible with edge devices is crucial, enabling the robotic platform to autonomously and in efficient perform inspection of vertical concrete surfaces with minimal computational resource consumption.

This paper introduces a novel defects detection system, CrackESS, a self-prompting crack segmentation system for edge devices, and demonstrated on open source datasets and a climbing robot system (see Fig. \ref{fig1}). Compared to pervious approaches, the main contributions of this paper can be summarized as follows:

1) We propose a novel self-prompting segmentation architecture for SHM of bridges. With this system, the robot has capabilities to perform efficient crack detection and segmentation on the captured images during the bridge surface inspection.

2) A fine-tuning approach for a lightweight SAM model is proposed, demonstrating that using the ConvLoRA method to fine-tune SAM can enhance the performance on downstream tasks while preserving its zero-shot capability.

3) Leveraging SAM's high sensitivity to prompts \cite{15}, a Crack Mask Refinement Module(CMRM) is proposed, which includes a specially designed keypoint extraction algorithm and an innovative CNN-based network that adaptively generates point prompts. Through this approach, the system achieves efficient and accurate crack segmentation while demonstrating exceptional performance on high-resolution crack images.

\section{Related Work}
Crack segmentation is a widely researched topic in the field of structural health monitoring (SHM). Due to the significant differences between crack images and natural images, achieving rapid and accurate crack segmentation could be a considerable challenge. CNN-based methods, such as CrackNet \cite{5}, SDDNet \cite{13}, CrackU-net \cite{6} and CrackW-net \cite{7}, have led to substantial progress in this field. However, considering the interference factors in practical implementations, the research on SAM-based crack segmentation methods shows a better performance, especially its zero-shot capability. Some recent studies, such as a fine-tuning SAM \cite{8} and CrackSAM \cite{9}, they have demonstrated high segmentation accuracy across various scenarios. However, their high computational resource requirements make them difficult to deploy on edge devices for efficient segmentation, despite the models performing remarkably in real-world tasks.

Fully training a large-scale foundational model is a resource-intensive process. In order to reduce training costs, a common approach is to employ PEFT methods for model fine-tuning. Ye et al. \cite{8} provide a comprehensive review of fine-tuning SAM through Parameter-Efficient Fine-Tuning (PEFT) approaches, categorizing them into two classes, fine-tuning lightweight mask decoder of SAM \cite{14}, and adding additional layers to the original SAM model, such as various lightweight adapters \cite{15} and \cite{16}, LoRA-based methods \cite{17}. The parallel structure design of LoRA and the implementation of the linear fine-tuning matrices ensure inference efficiency and could be easily integrated into the original projection matrix. Therefore, LoRA-based methods are widely used for fine-tuning SAM models in medical and concrete crack segmentation downstream tasks. 

Self-prompting is a common method that can automatically detect the prompts required by the SAM model from the input image. It replaces the manual input of prompts during the interactive procedures and enhances the level of automation in object segmentation substantially. Mobina et al. \cite{18} demonstrated that combining YOLOv8’s bounding box predictions with SAM achieves high segmentation accuracy while significantly reducing annotation time.

\section{Methodology}

\begin{figure}[!t]
	\centering
	\includegraphics[width=3.5in]{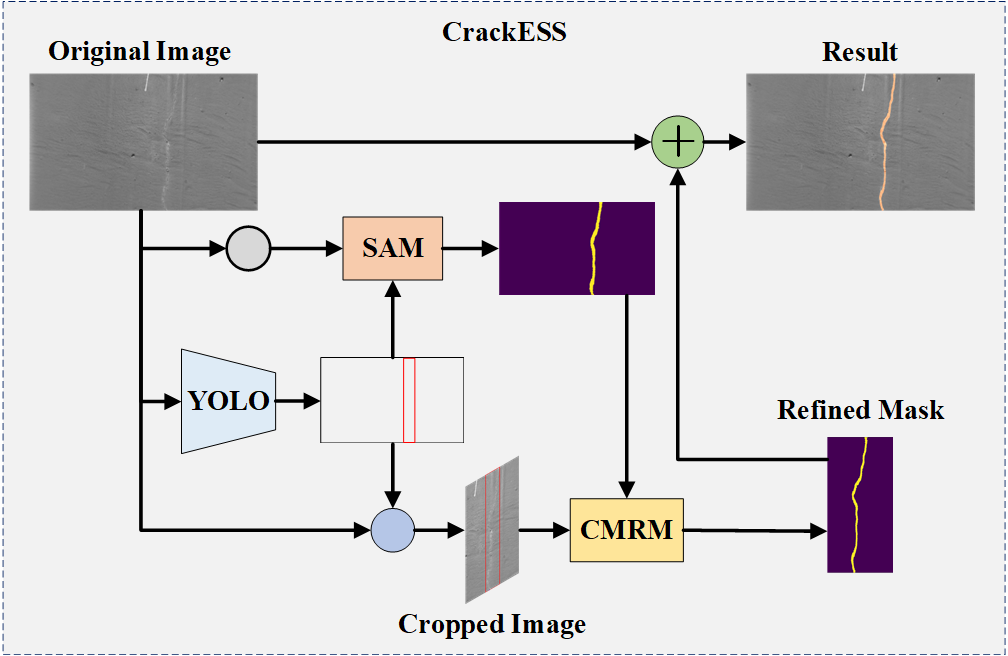}
	\caption{The framework of proposed CrackESS.}
	\label{fig2}
\end{figure}
\subsection{Self-prompting Crack Segmentation System}
\begin{figure*}[!t]
	\centering
	\includegraphics[width=7.1in]{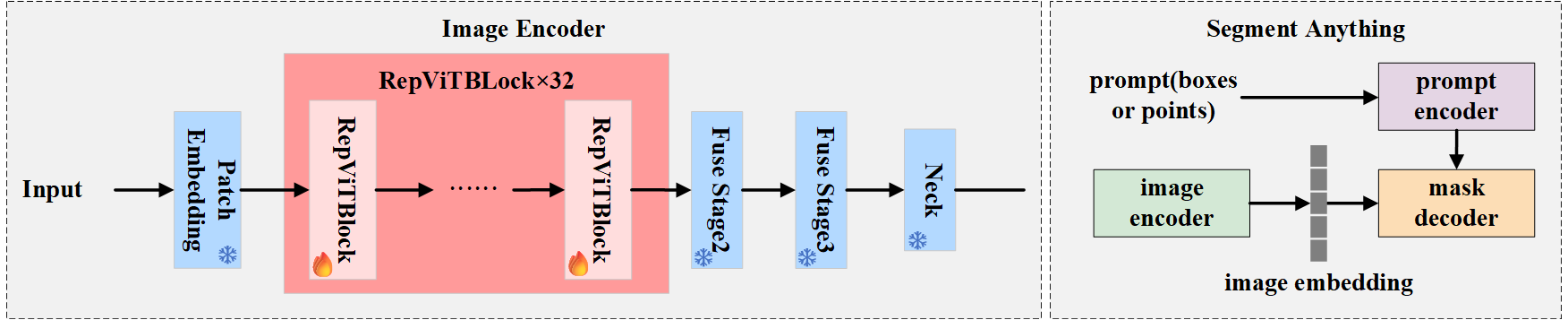}
	\caption{The image encoder block shows the architecture of EdgeSAM’s with trainable and frozen blocks. And the Segment Anything block shows the workflow of SAM.}
	\label{fig3}
\end{figure*}

\textit{1) Preliminary: } Segment Anything Model (SAM) is an efficient and promotable zero-shot model for image segmentation. As the overview shown in Fig. \ref{fig3}, the SAM model contains three modules: image encoder, prompt encoder and mask decoder \cite{19}. And the lightweight version of SAM models, such as FastSAM \cite{20}, EdgeSAM \cite{21} and MobileSAM \cite{22}, enable the possibility of efficient processing on edge devices for the original SAM model. EdgeSAM is a CNN-based architecture that is distilled from the original SAM by using a prompt-in-the-loop knowledge method. In this work, we select the EdgeSAM as the segmentation module, due to its remarkable performance \cite{21}.

\textit{2) Crack Segmentation Workflow:} The proposed self-prompting crack segmentation system integrates two fundamental models, YOLO and SAM, to effectively detect and segment cracks. The system is inspired by YOLO+SAM architecture \cite{23}, and the framework of the system is illustrated in Fig. \ref{fig2}. For instance, in the inspection tasks, the captured $3500\times2000$ image is first input to a pre-trained YOLOv8n model to generate box prompts. After preprocessing(cropping and resizing) the image, it is fed into a fine-tuned SAM model with the prompts for initial segmentation, then the resulting mask and the cropped image are processed by the CMRM module to obtain a refined mask. Finally, the optimized segmentation result is fused with the original image using a weighted combination. This architecture enables the crack detection system’s self-prompting functionality and enhances the performance of high-resolution segmentation results.
\subsection{The Fine-tuned Method}
\textit{1) Adaptation Method:} Considering that the image encoder of EdgeSAM is distilled from the original Vit-based SAM encoder, the common strategies for fine-tuning SAM by using LoRA may no longer be applicable. Inspired by ConvLoRA \cite{9} \cite{24}, we propose an effective fine-tuning method that enables it to achieve excellent performance in the crack segmentation. 

Specifically, in the fine-tuning of EdgeSAM image encoder, only the weights of the convolutional layers in the SEModule and Channel mixer of RepViTBlocks are trained, while the Patch Embedding, Fuse Stages 2 and 3, and Neck module are all frozen, as shown in Fig. \ref{fig3}. For RepViTBlocks adaptation, the ConvLoRA method is implemented in the fine-tuning process. ConvLoRA retains the core concept of LoRA, applying low-rank constraints through weight updates in convolutional layers, which makes it possible to achieve Parameter-Efficient Fine-Tuning (PEFT) of the neural network. Mathematically, the equation of ConvLoRA could be defined as follows:
\begin{align}
	\label{eq1}
	h=W_{0}x+W_{X}W_{Y}x
\end{align}
where ${{W}_{0}}\in {{\mathbb{R}}^{{{C}_{out}}\times {{C}_{in}}}}$,  ${{W}_{X}}\in {{\mathbb{R}}^{rank\times {{C}_{in}}}}$,  ${{W}_{Y}}\in {{\mathbb{R}}^{{{C}_{out}}\times rank}}$, and ${x}$ is the input tensor, ${rank}$ value is much smaller than the minimum of the input (${{C}_{in}}$) and output (${{C}_{out}}$) channel numbers. Besides, the weights of ${{W}_{X}}$ and ${{W}_{Y}}$ are initialized with a random Gaussian distribution and zero, respectively. In the training process, ${{W}_{0}}$ is frozen only the ${{W}_{X}}$ and ${{W}_{Y}}$ can receive updated gradients to train parameters. In the process of fine-tuning, the ConvLoRA adaptors are integrated into the conv2d layers of SEModule and Channel mixer. Through the experiments, the model shows the best performance of segmentation when ${rank=8}$.

\textit{2) Loss function:} We leverage DiceFocalLoss to address the crack dataset imbalance problems, thereby improving model performance. According to the equation proposed in [36], a typical form of Focal Loss can be defined as:
\begin{align}
	\label{eq2}
	{\mathcal{L}}_{FL} &= FL\left( {{p}_{t}} \right) = -{{\left( 1-{{p}_{t}} \right)}^{\gamma }}\log \left( {{p}_{t}} \right) 
\end{align}
\begin{align}
	\label{eq3}
	{{p}_{t}} &= \begin{cases}
		p,&{\text{when } y=1} \\ 
		{1-p,}&{\text{when } y=0} 
	\end{cases}
\end{align}

where $\gamma \ge 0$ is a tunable focusing parameter and set to 4 during training. 

For Dice Loss, a loss function commonly used in image segmentation tasks, the input P is the binarized prediction, and input G is the ground truth. The equation is as follows,
\begin{align}
	\label{eq4}
	{{\mathcal{L}}_{DL}}=DL\left( P,G \right)=1-~\frac{2\mathop{\sum }_{i=1}^{N}{{p}_{i}}{{g}_{i}}}{\mathop{\sum }_{i=1}^{N}p_{i}^{2}+\mathop{\sum}_{i=1}^{N}g_{i}^{2}}
\end{align}
where ${{p}_{i}}\in P~\left[ 0,1 \right]$ and ${{g}_{i}}\in G\left[ 0,1 \right]$.

DiceFocalLoss combines Focal Loss (Eq. (\ref{eq2}) and Dice Loss (Eq. (\ref{eq3})), with their respective functions shown below,
\begin{align}
	\label{eq5}
	{{\mathcal{L}}_{DFL}}=~{{\lambda }_{DL}}{{\mathcal{L}}_{DL}}+{{\lambda }_{FL}}{{\mathcal{L}}_{FL}}
\end{align}
where we set ${{\lambda }_{DL}}$ to 0.8, ${{\lambda }_{FL}}$ to 0.2.

\subsection{Crack Mask Refinement Module}
In the segmentation process, we noted that the fine-tuned module's result are potentially sensitive to the input image resolution, the location and number of point prompts, as well as the size and proportion of box prompts within the image. To improve the performance of the system, we propose a crack mask refinement module (CMRM) that can adaptively generate prompt points based on the input masks to enhance the performance of the system. In the following, we will provide the details of CMRM and the algorithm is illustrated in Algorithm\ref{alg1}. Notably, the extract-prompt-pts function in line 21 re-executes lines 6 to 19 of Algorithm 1.

The original image and the initial segmentation results generated by SAM are cropped and fed into the CMRM, with the region of interest is determined by enlarging the bounding box dimensions by 20 percent. To obtain a refined segmentation result, the first step is calculating an intersection map and a difference map that indicate the potential true positive (TP) and false positive (FP) regions. Notably, the map is from the input mask and a new mask,  which is based on re-positioned bounding boxes and the cropped image. Then the maps are eroded by using an adaptive-size elliptical structuring kernel to reduce or eliminate small and insignificant objects. In the case of a difference map, the algorithm extracts contours and selects those with an area larger than $m$ (e.g., $ m=50$, an empirical value). For each filtered contour, its maximum enclosing rectangle is computed, and Algorithm 2 extracts three middle points from each connected region in the maps, forming a set of positive and negative sample points based on the geometric properties of the bounding box. To improve segmentation accuracy, we employ two methods for point prompt selection. Firstly, CMRM picks a random point from each set. Secondly, after obtaining the initial points using the previous method, CMRM selects the point with the maximum Euclidean within their respective sets. In addition, to address the problem of extracted key points not being located in the nonzero regions of the maps, the find-centers function is designed in Algorithm 2. It follows these steps: identifying nonzero indices of the input array, segmenting them based on continuity, extracting the central index from each segment, and converting the extracted indices into coordinate points relative to the input image. 

\begin{algorithm}[!t]
	\caption{Crack Mask Refinement Module Algorithm}\label{alg1}
	\begin{algorithmic}[1]
		\renewcommand{\algorithmicrequire}{\textbf{Input:}}
		\renewcommand{\algorithmicensure}{\textbf{Output:}}
		\REQUIRE Bounding box $Boxes$, cropped image $I$, and the cropped segmentation result from the scaled image  $M$         
		\ENSURE Refined crack mask $R$  
		\STATE $R \leftarrow\{\}$
		\STATE {$features$, $M' \leftarrow$ SAM($I$, $Boxes$)}
		\STATE {$I_{diff}=M - M'$}
		\STATE {$K_{size} \leftarrow$ auto-kernel-size($I_{diff}$)}
		\STATE {$I_{erode} \leftarrow$ erode-with-ellipse ($I_{diff}, K_{size}$)}
		\STATE {$C\leftarrow$ find-contours ($I_{erode}$)}
		\STATE {$Points \leftarrow \{\}, Labels \leftarrow \{\}$}
		\IF{len($C$)$>0$} 
		\FOR{$i=1$ to len$(C)$}
		\IF{area($C$[$i$])$>m$}
		\STATE [$x, y, w, h$] $ \leftarrow$ bounding-box($C$[$i$])
		\STATE $Points[i] \leftarrow$ extract-points($I_{erode}, x, y, w, h$])
		\ENDIF
		\ENDFOR
		\STATE $P$ $ \leftarrow$ random($Points$])
		\STATE $D \leftarrow$ euclidean-distance($P$, $Points$)
		\STATE $P' \leftarrow$ $Points$[maximum-value-index($D$)]
		\STATE {$Points \leftarrow \{P, P'\}, Labels \leftarrow \{0, 0\}$}
		\ENDIF
		\STATE {$I_{inter}=M \cap M'$}
		\STATE $Points, Labels \leftarrow $extract-prompt-pts($I_{inter}$)
		\IF{len($P$)$>0$} 
		\STATE $R \leftarrow$ SAM($features, Boxes, Points, Labels$)
		\ELSE
		\STATE $R \leftarrow$ $M$
		\ENDIF
		\RETURN $R$
	\end{algorithmic}
\end{algorithm}

\begin{algorithm}[!t]
	\caption{Crack Mask Refinement Module Algorithm}\label{alg2}
	\begin{algorithmic}[1]
		\renewcommand{\algorithmicrequire}{\textbf{Input:}}
		\renewcommand{\algorithmicensure}{\textbf{Output:}}
		\REQUIRE difference image or intersection image $I$, x and y value of the top left vertex coordinate of the rectangle, and the rectangle's width $w$ and height $h$        
		\ENSURE Refined crack mask $R$  
		\STATE $P \leftarrow \{\}$
		\IF{$w > h$} 
		\STATE $l \leftarrow w // n$
		\FOR{$i = 1$ to $n - 1$}
		\STATE $C \leftarrow $find-centers($I[y : y + h, x + i \cdot l]$)%
		\STATE $P \leftarrow P \cup \{(x + i \cdot l, y + C[\text{len}(C) // 2])\}$
		\ENDFOR
		\ELSE
		\STATE $l \leftarrow h // n$
		\FOR{$i = 1$ to $n - 1$}
		\STATE $C \leftarrow $find-centers($I[y + i \cdot l, x : x + w]$)
		\STATE $P \leftarrow P \cup \{(x +C[\text{len}(C) // 2], y + i \cdot l)\}$
		\ENDFOR
		\ENDIF
		\RETURN $P$
	\end{algorithmic}
\end{algorithm}

Specifically,  we propose a lightweight CNN model to predict the kernel size of the elliptical structuring element for subsequent morphological processing. The model's backbone consists of a convolutional layer and StarNet blocks(the depth parameter of StarNet to $[1,1,1,1]$) for producing features from the input maps. These features are then processed through a $3\times3$ adaptive average pooling layer before being fed into the head section, which comprises three fully connected layers. In the training process, we establish a set of odd integers $K$ , for each kernel size from $K$ is applied to process the difference image $I_{diff}$ for the target re-segmentation, and the resulting refined mask evaluated  against the ground truth to calculate their $IoU$ scores, thereby constructing the corresponding set $S$, 
\begin{align}
	\label{eq6}
	K = \{3, \ldots , 2n+1\}
\end{align}
\begin{align}
	\label{eq7}
	S=\{~\frac{M \cap M'_{0}}{M \cup M'_{0}},\ldots,~\frac{M \cap M'_{n}}{M \cup M'_{n}}\}
\end{align}
where the set $K$ contains odd numbers ranging from 1 to 2n+1. According to eq6 and eq7, the element $S_n$ of $S$ is generated from kernel size $K_n$ in $K$. For instance, when $n=1$, the element $M_{1}'$ of set $S$ corresponds to the refined mask generated with a kernel size of 3. Therefore, the CNN model uses the following loss function in terms of huber loss ($\delta=0.3$) on kernel sizes
\begin{align}
	\label{eq8}
	{{L}_{\delta}(G,k)} &= \begin{cases}
		~\frac{1}{2}{(G-k)}^{2}),&{|G-k|\leq\delta} \\ 
		\delta|G-k|-~\frac{1}{2}{\delta}^2,&{|G-k|>\delta}
	\end{cases}
\end{align}
where $k$ is produced by the model and G represents the mapping of the maximum $IoU$ score from the set $S$ to the corresponding kernel size in the set $K$, as defined by the following equation:
\begin{align}
	\label{eq9}
	G = K[\arg\max_{n} S_n]
\end{align}
Notably, we set parameters $n=15$ enabling the model to predict fifteen odd kernel sizes ranging from 3 to 31.

The CMRM is a simple but effective module that can improve the performance of crack segmentation, particularly when processing high-resolution images.  In the following, we detail the performance of the proposed method by using two different high-resolution datasets.

\section{Experiments and Results}
\subsection{Dataset}
A dataset compiled by Khanhha \cite{25} is utilized for fine-tuning the image encoder of SAM model. This dataset integrates seven different datasets, comprising 9,603 training and 1,695 testing images, along with their corresponding GT masks, all with a resolution of 448×448 pixels. During the training and testing, the $448\times448$ resolution Crack500\cite{26} subset was completely removed, and instead, its high-resolution $2560\times1440$ version was used for evaluating the CraskESS's performance. Additionally, we constructed a dataset containing 4,517 images with a resolution of $3500\times2000$ pixels from concrete bridge pier inspection tasks.Within this dataset includes 39 images with cracks, and CrackCR consists of these images and their corresponding GT masks. 

\begin{figure}[!t]
	\centering
	\includegraphics[width=3.5in]{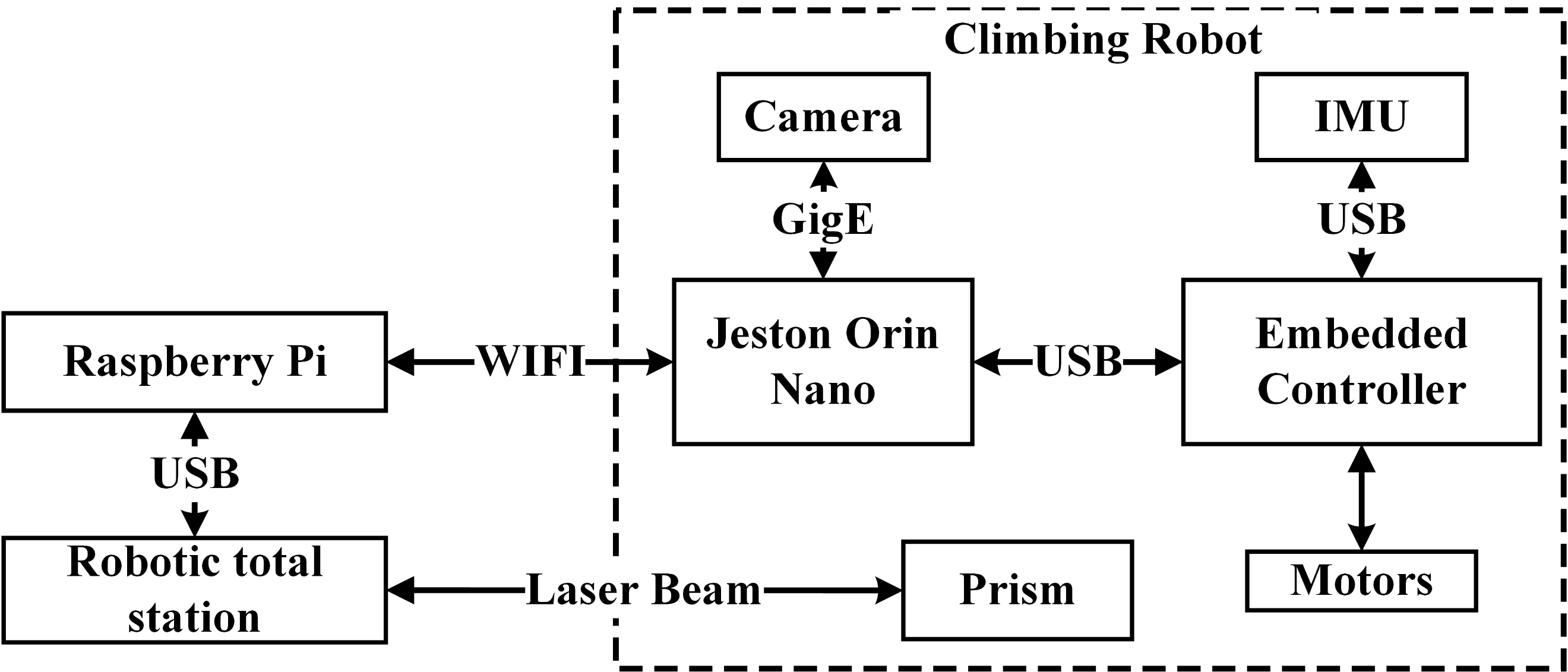}
	\caption{The climbing robot system using robotic total station and trackable prism.}
	\label{fig4}
\end{figure}
\subsection{Climbing Robot System}
The CrackESS system was validated through real-world tasks using a climbing robot equipped for autonomous image collection on concrete bridge surfaces. The robot utilizes a pneumatic adhesion mechanism  \cite{11} for stable climbing, supported by an electrical vacuum generator and a differential drive system. The image acquisition module consists of an industrial camera ($3500\times2000$) mounted on the robot, which captures high-resolution images during bridge inspections. During the inspection, the robot utilizes a robotic TS (total station) and multiple sensors, including IMU and encoders, for precise localization and navigation, achieving automatic image acquisition through full-coverage path planning of the target work area. The architecture of the climbing robot system is shown in Fig. \ref{fig4}. Through this design, the robot achieves automated image-based inspection of vertical concrete bridge surfaces. The collected data supports the validation of the CrackESS in real-world scenarios.

\subsection{Implementation Details}
For the evaluation of the fine-tuned model, the image encoder of the model is processed using the ConvLoRA method mentioned in Section III, then trained for 100 epochs using the Adam optimizer with 1e-4 learning rate (no weight decay), and DiceFocalLoss is used as the loss function. In addition, the batch size is set to 4 due to the GPU's performance limitation. Refer to the metrics hired in [2], precision, recall, IoU, and dice score are used to evaluate the performance of our fine-tuned model with other SOTA models on the Khanhha's test dataset. Moreover, CrackESS is assessed on the Crack500 and CrackCR datasets using the same evaluation approach to validate the effectiveness of CMRM. Notably, the boxes inferred by YOLOv8n are replaced with the bounding rectangles of the ground truth masks to minimize the impact of prompt fluctuations in the above experiments. 

We implemented the proposed CracESS on the climbing robot(Nvidia Orin Nano 8GB) platform and accelerated it using TensorRT, enabling the system could detect and segment cracks in efficient during the task. During the inspection tasks, the input images have a resolution of $3500 \times 2000$, and the collected data, along with the corresponding computation results, will be saved locally on the robot..
\begin{table}[!t]
	\begin{center}
		\centering
		\tabcolsep=0.00cm
		\renewcommand{\arraystretch}{1.5}
		\caption{Performance comparison of different models on Khanhha's dataset.}
		\label{tab1}
		\begin{tabular}{ m{1.8cm}<{\centering} |m{1.35cm}<{\centering}  m{1.35cm}<{\centering} m{1.35cm}<{\centering} m{1.35cm}<{\centering} | m{1.35cm}<{\centering }}
			\Xhline{2pt}
			\bf Method  &\bf Precision & \bf	Recall & \bf IoU & \bf Dice & \bf	FPS \\
			\Xcline{1-1}{0.4pt}
			\Xhline{1pt}
			CrackW-net	 				&32.6\%  &66.4\%	&25.8\%	&38.9\%	&8.237 \\
			EdgeSAM		  				&18.4\%  &76.5\%	&15.8\%	&24.9\%	&63.414 \\ 
			CrackSAM	 	 			&74.7\%  &\bf77.2\%	&\bf60.9\%	&\bf74.6\%	&14.579 \\ 
			Fine-tuned model (r=2)		&75.7\%	&70.3\%	&55.8\%	&70.5\%	&63.380 \\
			Fine-tuned model (r=4)		&74.5\%	&72.0\%	&56.3\%	&71.0\%	&63.223 \\
			Fine-tuned model (r=8)     	&74.3\%	&\bf73.8\%	&\bf56.9\%	&\bf71.4\%	&\bf62.391 \\ 
			Fine-tuned model (r=16)	    &\bf76.3\%	&69.7\%	&56.0\%	&70.7\%	&61.497 \\ 
			\Xhline{2pt}
		\end{tabular} 
	\end{center}
\end{table}
\begin{table}[!t]
	\begin{center}
		\centering
		\tabcolsep=0.00cm
		\renewcommand{\arraystretch}{1.5}
		\caption{Performance comparison of proposed system on Crack500 dataset.}
		\label{tab2}
		\begin{tabular}{ m{1.8cm}<{\centering} |m{1.35cm}<{\centering}| m{1.35cm}<{\centering}  m{1.05cm}<{\centering} m{1.05cm}<{\centering} m{1.05cm}<{\centering} | m{1.05cm}<{\centering }}
			\Xhline{2pt}
			\bf Method  &\bf Dataset &\bf Precision & \bf	Recall & \bf IoU & \bf Dice & \bf	FPS \\
			\Xcline{1-1}{0.4pt}
			\Xhline{1pt}
			Fine-tuned model (r=8)		        &Crack500 &50.5\%	&69.3\%	&37.8\%	&52.7\%	&32.415 \\
			\makecell{CrackESS \\ (2-point)}	&Crack500 &51.7\%  &69.9\%	&39.5\%	&54.4\%	&13.804 \\ 
			\makecell{CrackESS \\ (4-point)}	&Crack500 &51.7\%  &\bf70.5\%	&\bf40.7\%	&\bf 56.1\%	&13.803 \\ 
			\makecell{CrackESS \\ (6-point)}	&Crack500 &\bf52.0\%  &69.5\%	&40.3\%	&55.7\%	&13.811 \\ 
			\Xhline{2pt}
		\end{tabular} 
	\end{center}
\end{table}

\begin{table}[!t]
	\begin{center}
		\centering
		\tabcolsep=0.00cm
		\renewcommand{\arraystretch}{1.5}
		\caption{Performance comparison of proposed system and CrackSAM on CrackCR dataset}
		\label{tab3}
		\begin{tabular}{ m{1.8cm}<{\centering} |m{1.35cm}<{\centering}| m{1.35cm}<{\centering}  m{1.05cm}<{\centering} m{1.05cm}<{\centering} m{1.05cm}<{\centering} | m{1.05cm}<{\centering }}
			\Xhline{2pt}
			\bf Method  &\bf Dataset &\bf Precision & \bf	Recall & \bf IoU & \bf Dice & \bf	FPS \\
			\Xcline{1-1}{0.4pt}
			\Xhline{1pt}
			CrackSAM 							&CrackCR	&\bf79.2\%  	&40.6\%  	&36.9\%		&48.2\%		&2.511	\\
			Fine-tuned model (r=8)				&CrackCR 	&58.2\%	 	&\bf 78.8\%	&48.0\%		&63.4\%		&24.213 \\
			\makecell{CrackESS \\ (2-point)}	&CrackCR 	&62.6\%  		&74.0\%		&48.7\%		&64.5\%		&10.859 \\ 
			\makecell{CrackESS \\ (4-point)}	&CrackCR 	&60.7\%  		& 76.4\%		&\bf49.6\%	&\bf 64.9\%	& 10.857\\ 
			\makecell{CrackESS \\ (6-point)}	&CrackCR 	&59.1\%  		&76.4\%		&48.8\%		&64.3\%		& 10.860\\ 
			\Xhline{2pt}
		\end{tabular} 
	\end{center}
\end{table}
\subsection{Experimental Results}
\textit{1) Segmentation performance:}  TABLE \ref{tab1} shows the segmentation performance of our fine-tuned SAM model compared to state-of-the-art (SOTA) methods, evaluated on Khanhha’s test dataset (1,695 images). Compared with CNN-based CrackW-net, the fine-tuned model achieves significantly improved segmentation accuracy while maintaining superior inference speed. However, CrackSAM outperforms our model in IoU and Dice score by approximately 0.04. We also experimented with different rank values ($r=2,4,8,16$), and the model achieved the best segmentation performance with a dice score of 71.4\% when the rank was set to 8.  Additionally, the results of FPS in TABLE \ref{tab1} show our model is approximately 4.3$\times$ faster than CrackSAM and 7.7 $\times$ faster than CrackW-net. The above comparison demonstrates that our model ensures the accuracy of crack segmentation while also achieving remarkable inference speed. 

\begin{figure*}
	\begin{center}
		\centering
		\tabcolsep=0.00cm
		\renewcommand{\arraystretch}{0.8}
		\begin{tabular}{m{1.5cm}<{\centering} | m{2cm}<{\centering} m{2cm}<{\centering} m{2cm}<{\centering}  m{2cm}<{\centering}  m{2cm}<{\centering}  m{2cm}<{\centering}  m{2cm}<{\centering}  m{2cm}<{\centering}}
			\toprule
			\bf Image& \includegraphics[width=1.8cm]{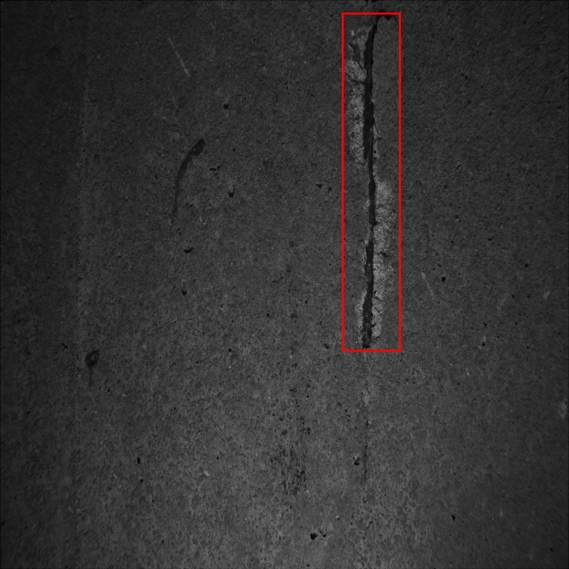}      & \includegraphics[width=1.8cm]{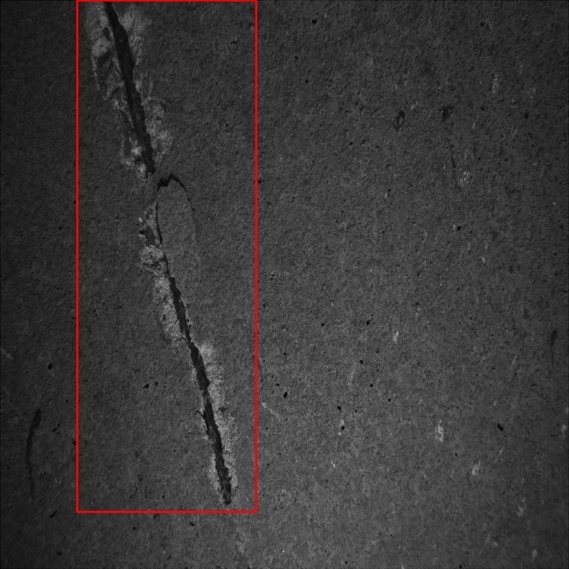}   & \includegraphics[width=1.8cm]{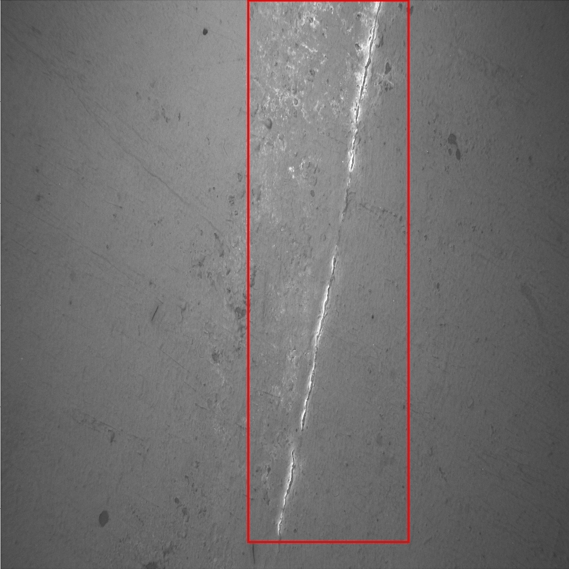} & \includegraphics[width=1.8cm]{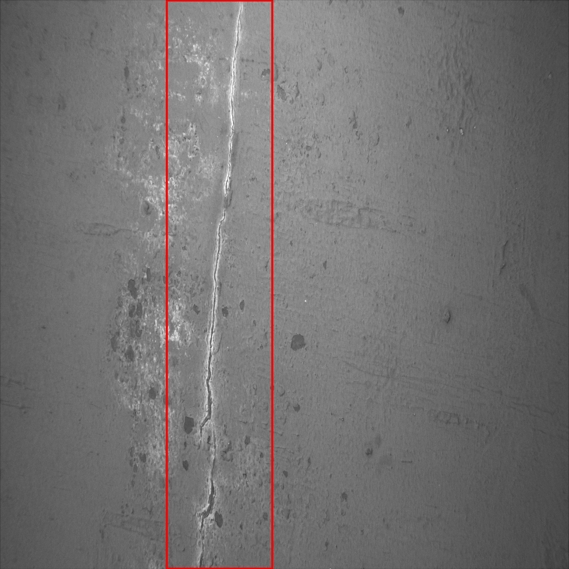} & \includegraphics[width=1.8cm]{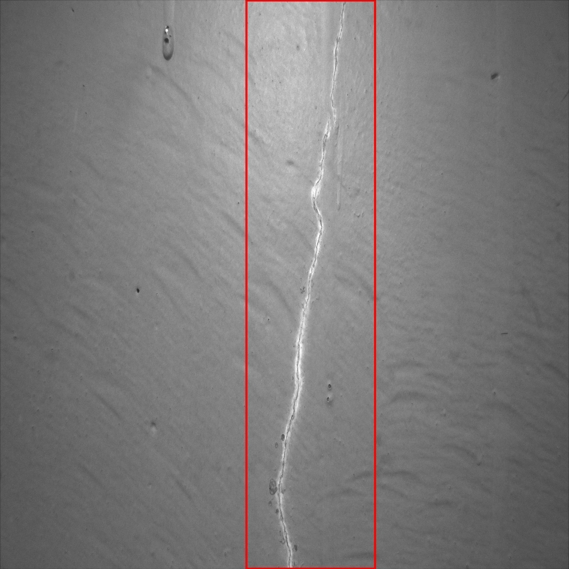} & \includegraphics[width=1.8cm]{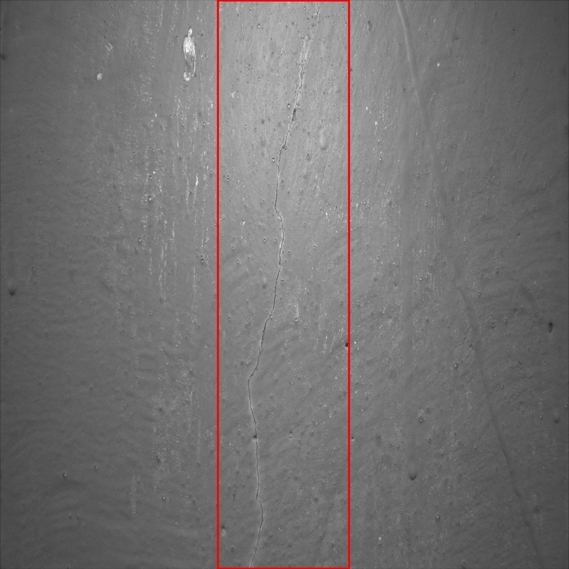}& \includegraphics[width=1.8cm]{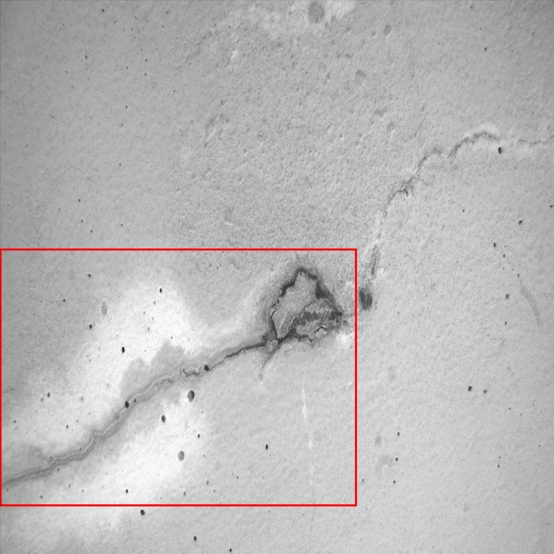}& \includegraphics[width=1.8cm]{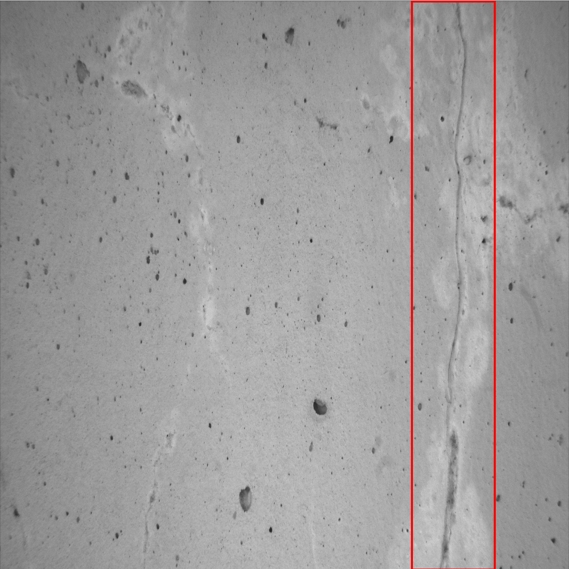}            \\ 
			
			\bf Ground Truth& \includegraphics[width=1.8cm]{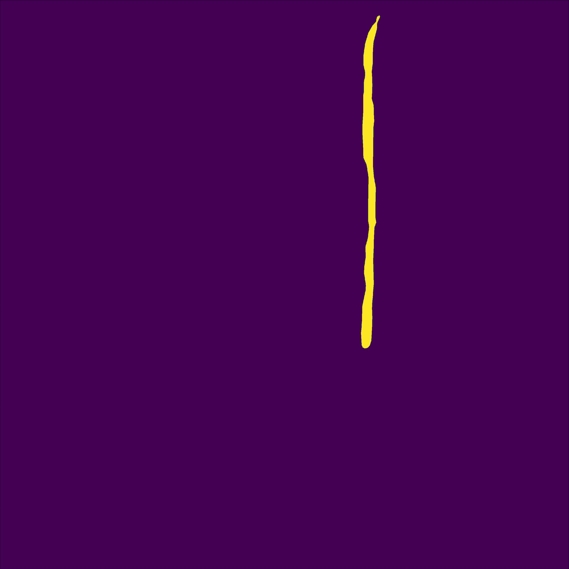}      & \includegraphics[width=1.8cm]{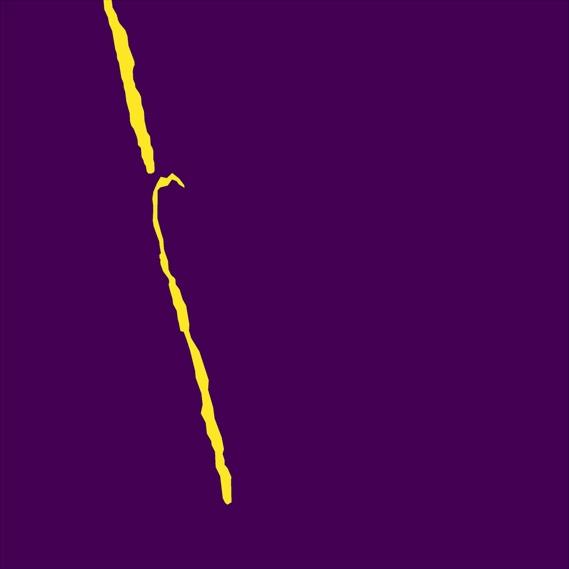}   & \includegraphics[width=1.8cm]{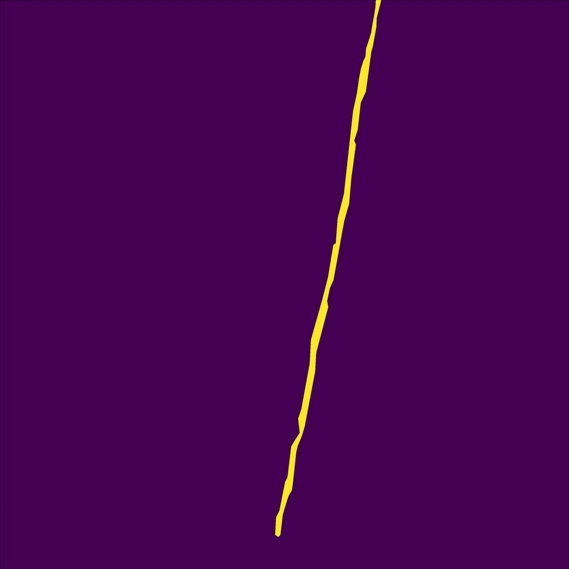} & \includegraphics[width=1.8cm]{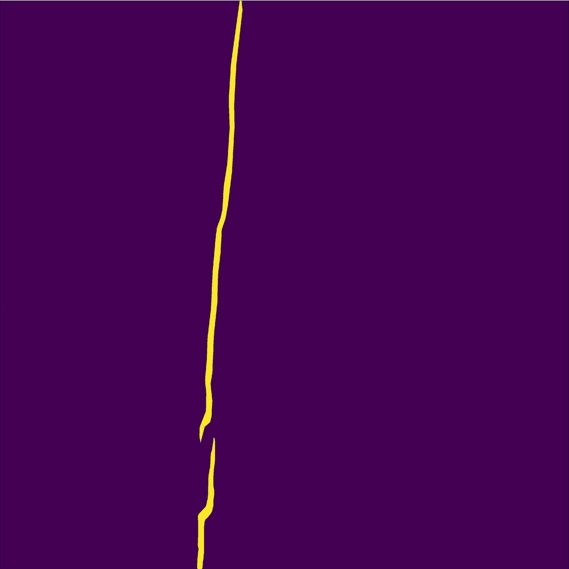} & \includegraphics[width=1.8cm]{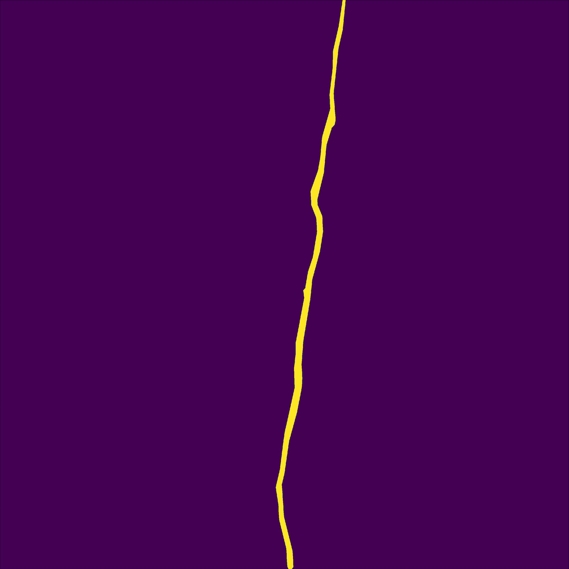} & \includegraphics[width=1.8cm]{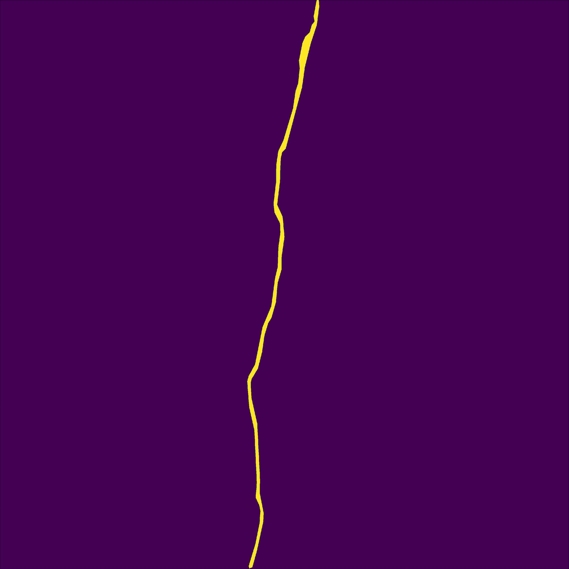}& \includegraphics[width=1.8cm]{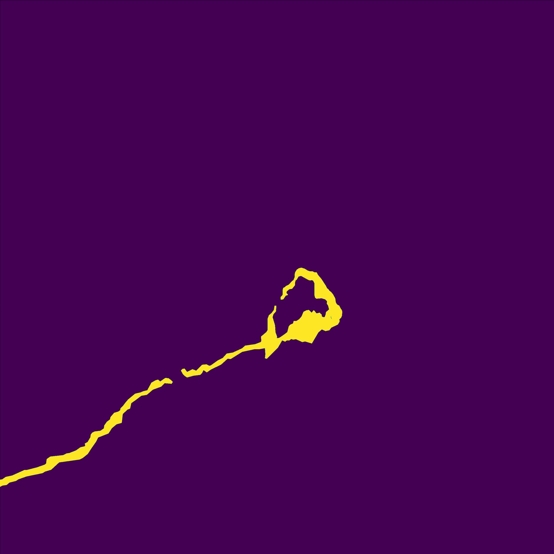}& \includegraphics[width=1.8cm]{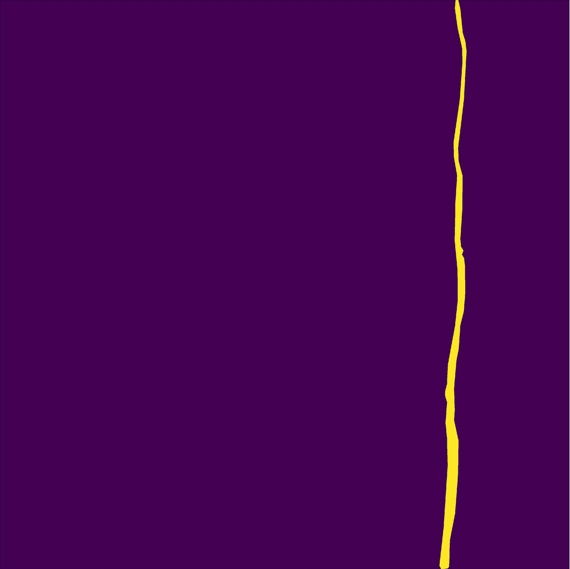}            \\ 
			
			\bf Initial Result& \includegraphics[width=1.8cm]{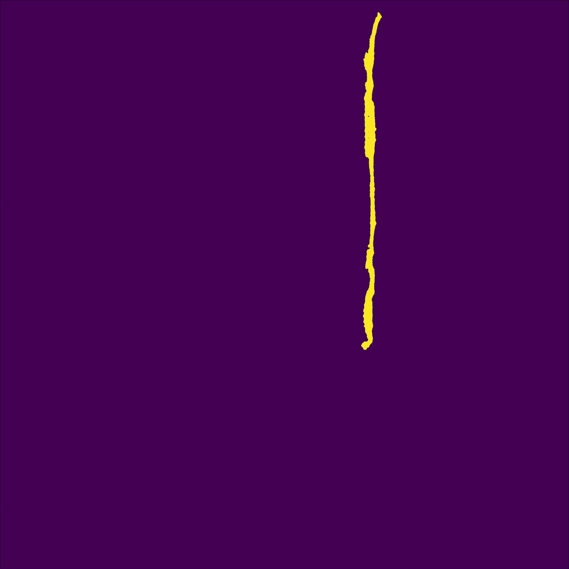}      & \includegraphics[width=1.8cm]{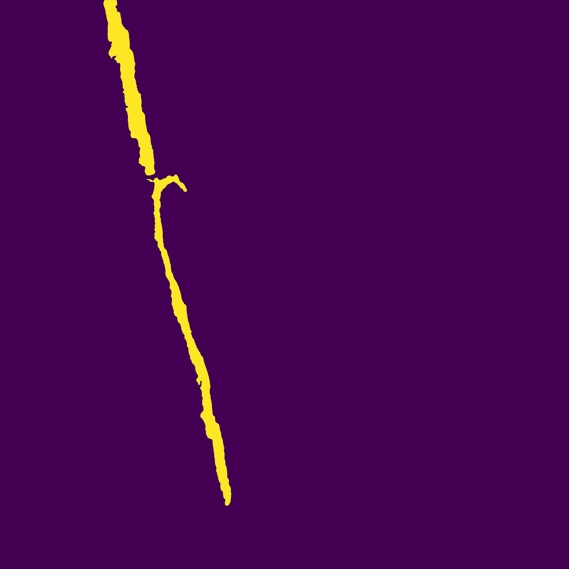}   & \includegraphics[width=1.8cm]{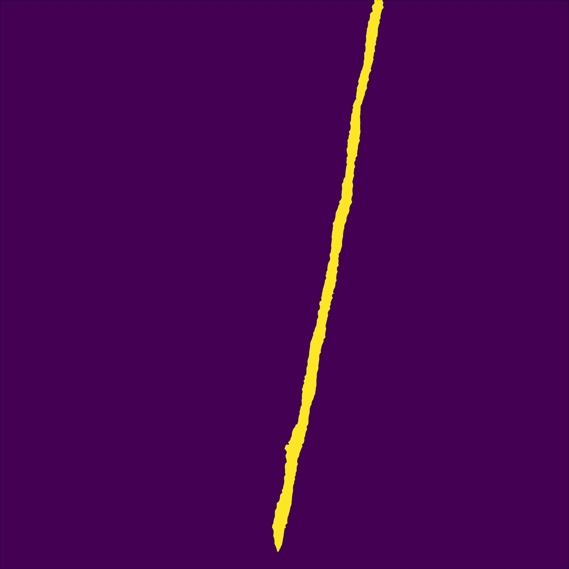} & \includegraphics[width=1.8cm]{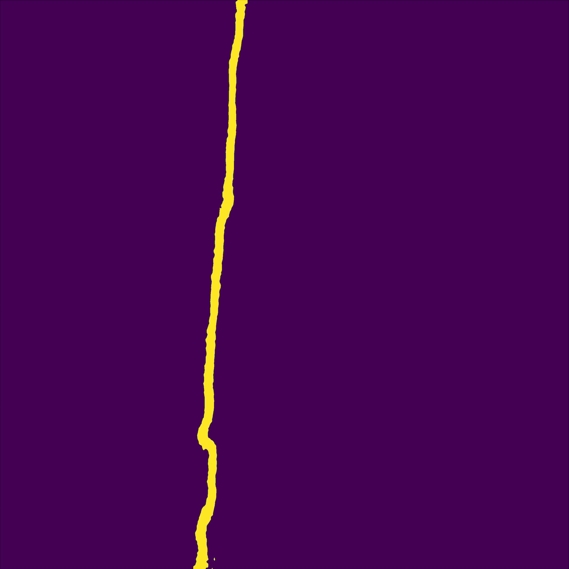} & \includegraphics[width=1.8cm]{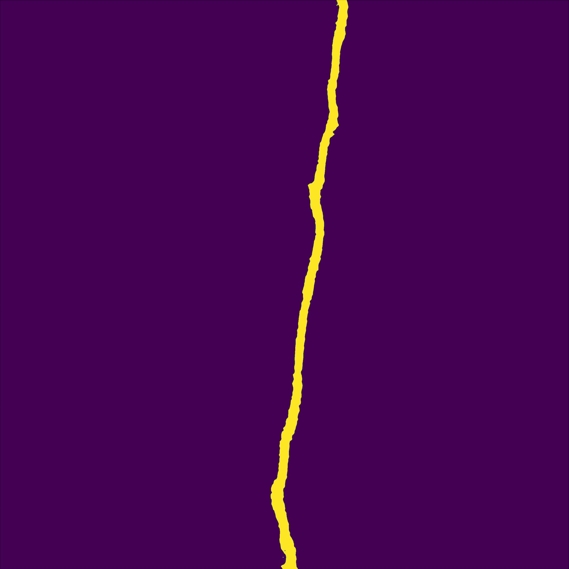} & \includegraphics[width=1.8cm]{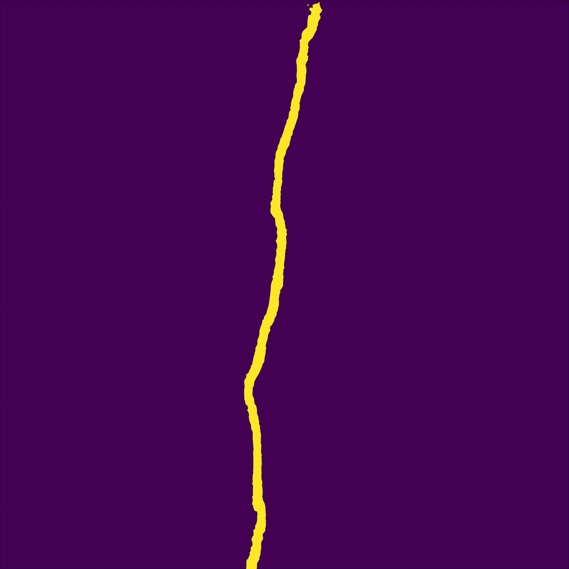}& \includegraphics[width=1.8cm]{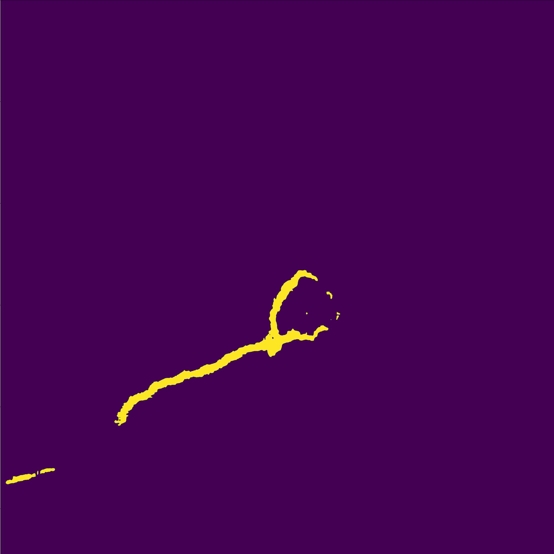}& \includegraphics[width=1.8cm]{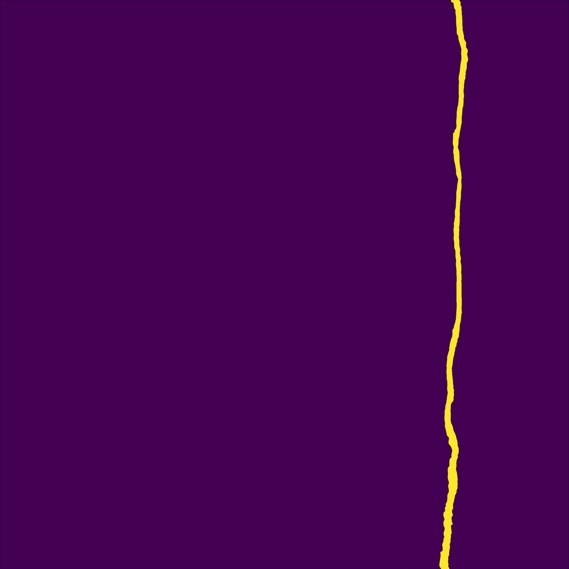}            \\ 
			
			\bf Point Prompts& \includegraphics[width=1.8cm]{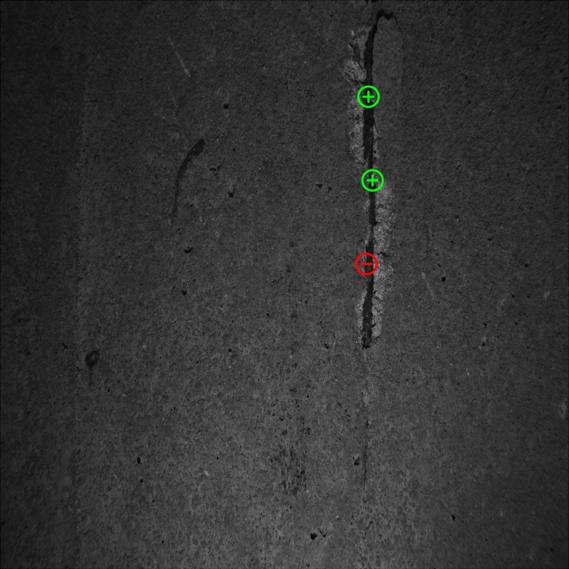}     & \includegraphics[width=1.8cm]{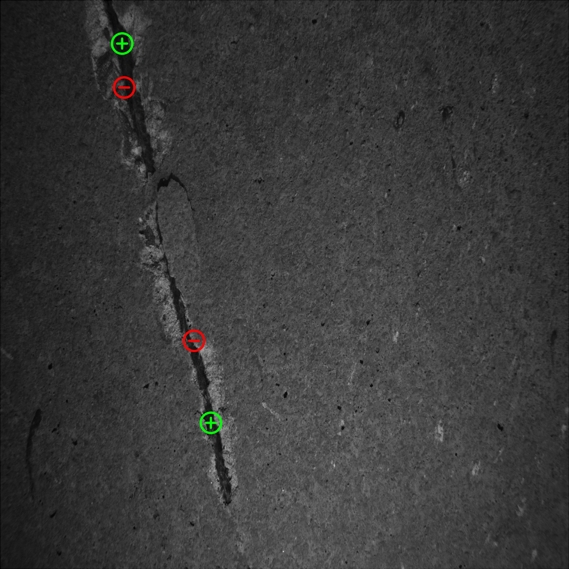}   & \includegraphics[width=1.8cm]{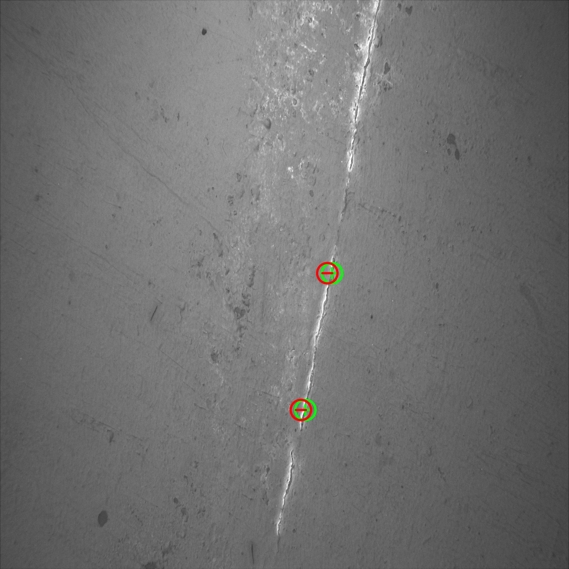} & \includegraphics[width=1.8cm]{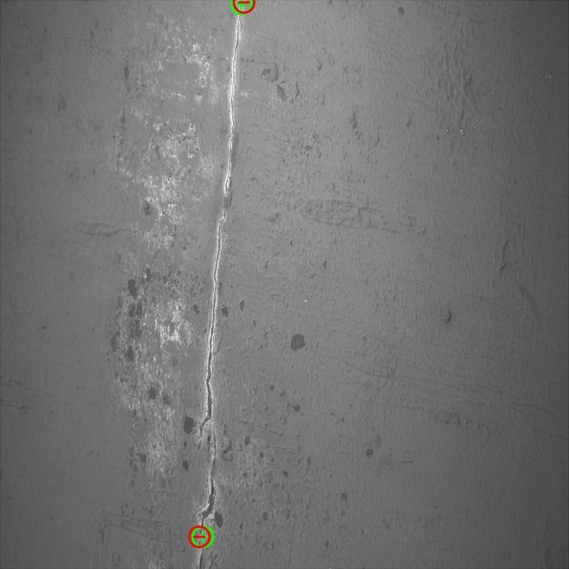} & \includegraphics[width=1.8cm]{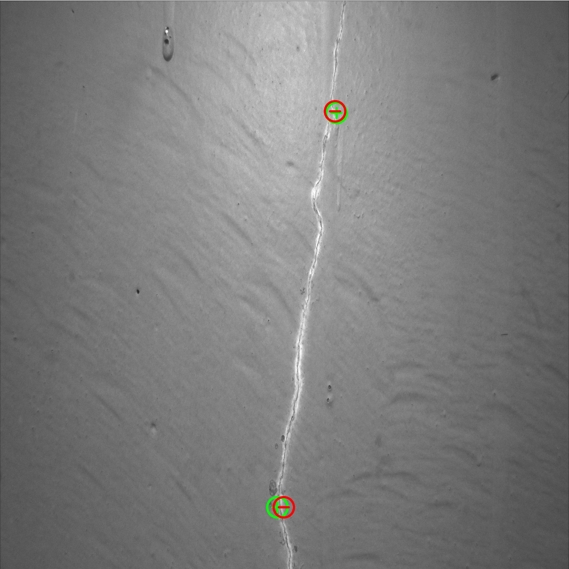} & \includegraphics[width=1.8cm]{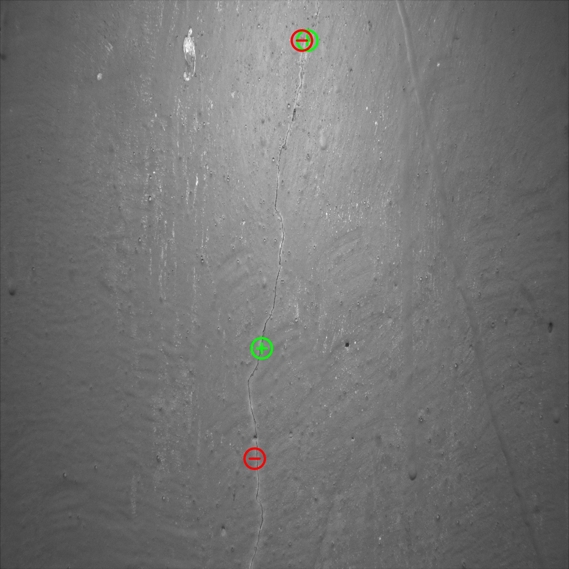}& \includegraphics[width=1.8cm]{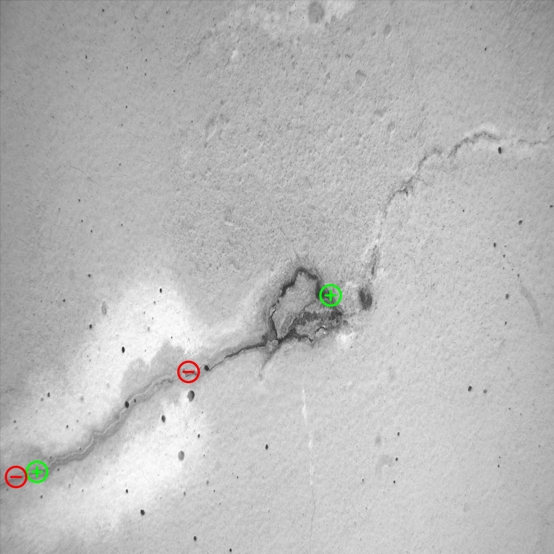}& \includegraphics[width=1.8cm]{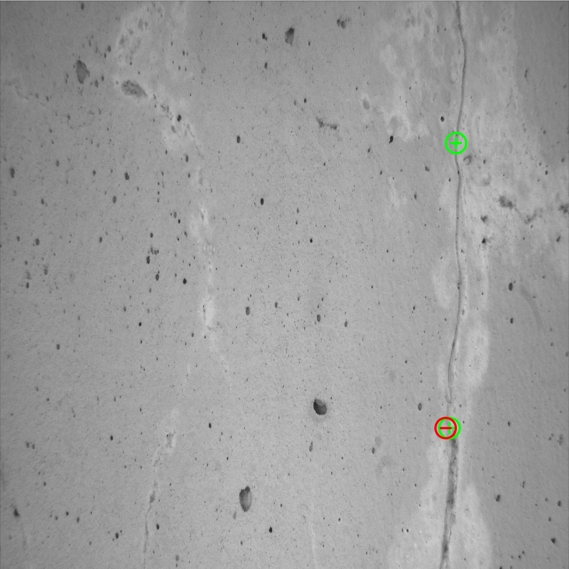}            \\
			
			\bf Refined Result& \includegraphics[width=1.8cm]{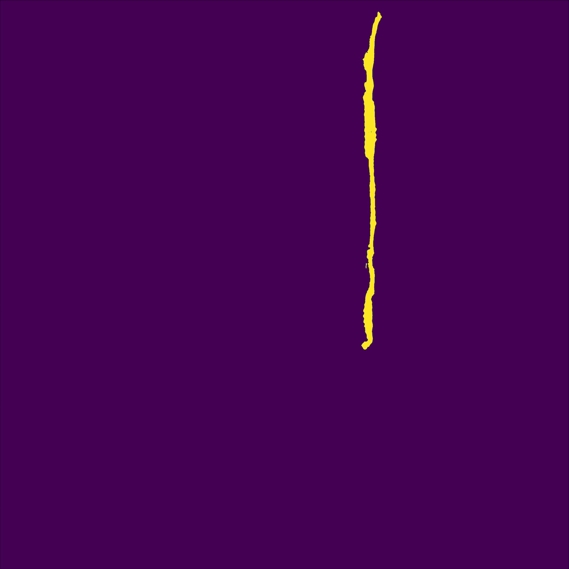}     & \includegraphics[width=1.8cm]{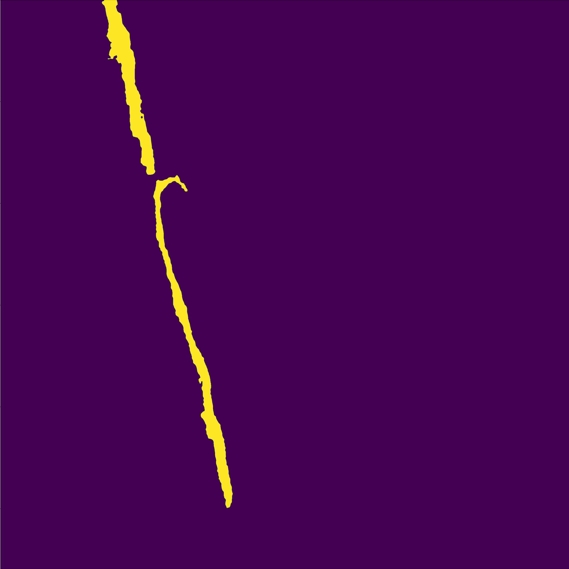}   & \includegraphics[width=1.8cm]{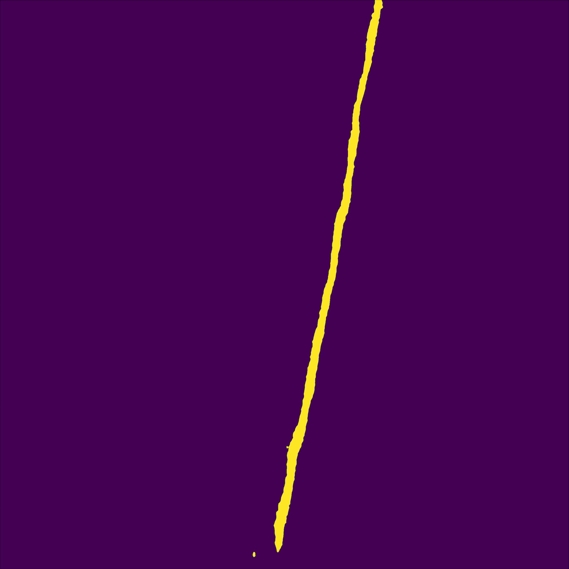} & \includegraphics[width=1.8cm]{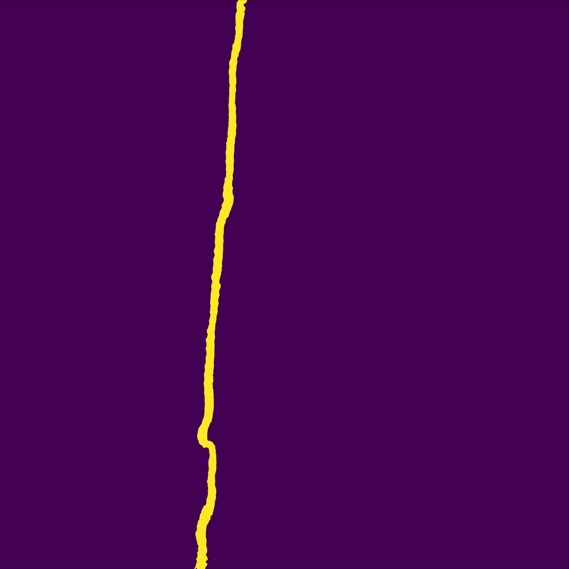} & \includegraphics[width=1.8cm]{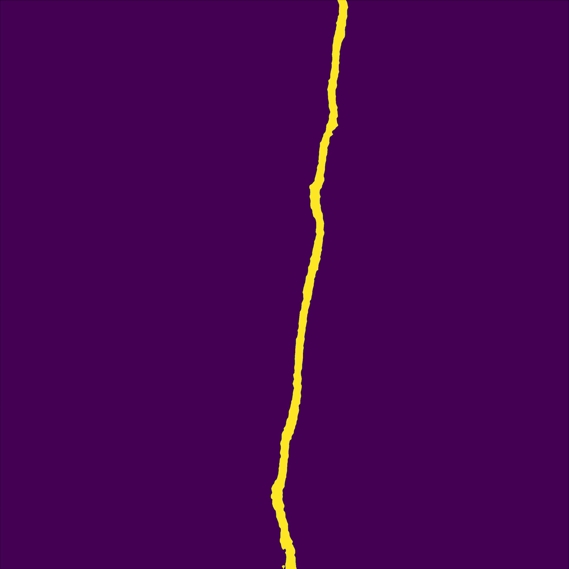} & \includegraphics[width=1.8cm]{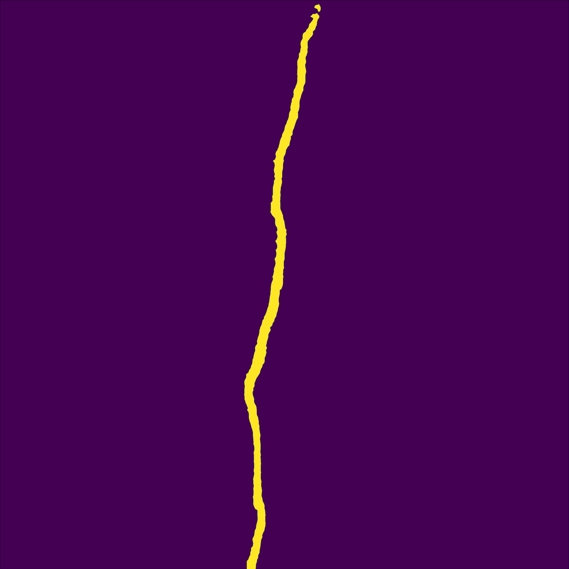}& \includegraphics[width=1.8cm]{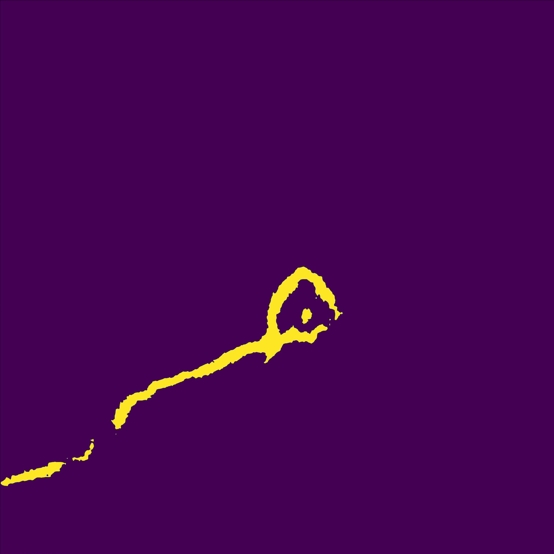}& \includegraphics[width=1.8cm]{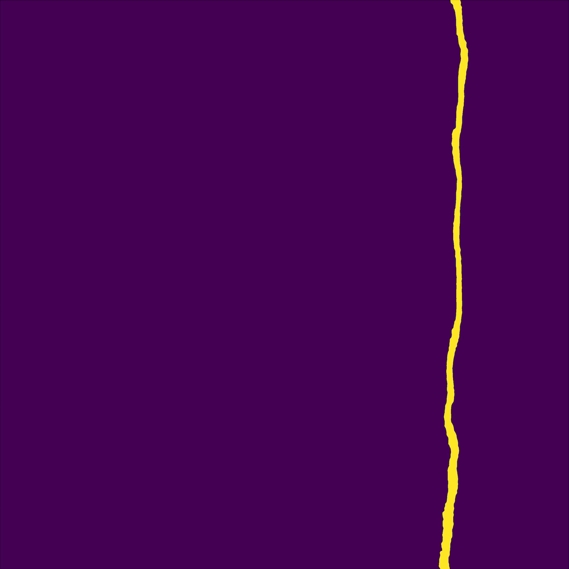}            \\
			\toprule
			\bf Dice1 &  78.2\%  &  83.0\% &  72.2\% &  75.0\% &  75.7\%&  59.9\%&  54.0\% &  80.4\%          \\
			\toprule
			\bf Dice2 &  80.2\%  &  86.1\% &  79.5\% &  77.7\% &  79.6\%&  62.2\%&  63.1\% &  81.2\%          \\
			\toprule
		\end{tabular}
	\end{center}
	\caption{CrackESS visualization results of each module in bridge inspection experiment. The green circles with plus signs in the Point Prompts row represent positive point prompts, while the red circles with minus signs indicate negative point prompts.}
	\label{fig5}
\end{figure*}
TABLE \ref{tab2} summarizes the performance of our fine-tuned model and CrackESS variants on the Crack500 datasets. As shown in TABLE \ref{tab2}, CrackESS(4-point) achieved the highest Dice score of 56.1\%, improving segmentation accuracy by 3.4\%  compared to the fine-tuned model. Similarly, TABLE \ref{tab3} performs metrics of CrackESS on CrackCR, the 4-point approach achieved the best Dice score of 64.9\%, outperforming other prompt configurations. The 4-point (two positive and two negative prompt points) demonstrated optimal segmentation accuracy, while the 2-point and 6-point methods resulted in slightly lower performance. Notably, CrackESS is approximately four times faster in inference speed than CrackSAM and achieves superior performance on the CrackCR dataset.

\begin{table}
	\begin{center}
		\centering
		\tabcolsep=0.00cm
		\renewcommand{\arraystretch}{1.5}
		\caption{Performance comparison on different networks.}
		\label{tab4}
		\begin{tabular}{m{1.86cm}<{\centering} m{1.72cm}<{\centering} m{1.72cm}<{\centering} m{1.72cm}<{\centering}  m{1.46cm}<{\centering}}
			\toprule
			\bf Networks & \bf Total Params & \bf	Trainable Params & \bf Input and Size & \bf Params Size  \\ \hline
			CrackESS &  5517552 & 	46992 &  12.00MB &  21.05MB  \\ 
			\toprule
		\end{tabular}
	\end{center}
\end{table}


\textit{2) The System’s Performance on Climbing Robot:} In order to assess the practical application of CrackESS, we conducted automated inspections on two concrete bridge piers. Fig. \ref{fig5} presents the results of each module in CrackESS during the inspection tasks, along with a comparison of Dice scores between the initial and refined segmentation results calculated against the ground truth. The red boxes in the first row represent the bounding boxes generated by YOLOv8, and the second row shos the GT mask, obtained using offline tools and manual adjustments. The third row displays the results derived from the geometric information of the boxes, achieved by cropping and segmenting the original image. The prompts generated by CMRM are shown in the forth row, with green and red circle are representing positive and negative points, respectively. 

In the practical experiments, TABLE \ref{tab4} shows the parameters of the fine-tuned SAM model in CrackESS. By comparing the initial and refined results, it visually demonstrates that the CMRM has refined the initial segmentation results, and improving precision and Dice score. As shown in the 7th of Fig. \ref{fig5},  the initial result is deficient in detail, with some missing regions compared to the ground truth (GT), resulting in a Dice score of 54.0\%. After the refinement processed by CMRM, the refined result provides a more detailed and comprehensive segmentation, leading to a 9\% improvement in the Dice score.

\section{Conclusion}
This paper proposes a novel CrackESS for efficient crack detection and segmentation on edge devices. It is a self-prompting crack segmentation system that integrates YOLOv8, a light-weight fine-tined SAM model and CMRM module. CrackESS leverages the zero-shot capability of the SAM model, offering greater adaptability compared to CNN-based methods. Additionally, by employing a specially designed PEFT method to fine-tune SAM, our system achieves an inference speed at least four times faster than existing SAM-based methods while maintaining comparable accuracy. The Crack Mask Refinement Module (CMRM) dynamically process the maps and extracts point prompts to enhance crack segmentation performance, especially for high-resolution defect images. We also demonstrate the integration of our system into a climbing robot system equipped with localization and navigation capabilities, and the effectiveness of our approach is validated through the experiments. The paper provides insights into the development of an efficient crack detection and segmentation system for edge devices. In the future, we will further enhance the system's performance form the perspectives of accuracy and robustness while expanding its capabilities for more precise analysis and assessment of defects in vertical infrastructure.

\end{document}